\documentclass{ecai}
\usepackage{times}
\usepackage{graphicx}
\usepackage{latexsym}

\usepackage{subfigure}
\usepackage{algorithm}
\usepackage{algorithmic}
\usepackage{amsfonts}
\usepackage{amsmath}
\usepackage{multicol}
\usepackage{multirow}
\usepackage{url} 
\usepackage{setspace}
\RequirePackage{color}

\newcommand{\etal}{\textit{et al}.}
\newcommand{\etc}{\textit{etc}. }

\newcommand{\eg}{\textit{e}.\textit{g}.}

\begin{document}
	
	\title{An Efficient Agreement Mechanism in CapsNets By Pairwise Product}
	
	\author{Lei Zhao\institute{TalkingData,
			China, email: bhneo@126.com}
		\and Xiaohui Wang\institute{TalkingData,
			China, email: xiaohui96@gmail.com}
		\and Lei Huang\institute{Inception Institute of Artificial Intelligence,
			UAE, email: huanglei36060520@gmail.com} }
	
	\maketitle
	\bibliographystyle{ecai}
	
	\begin{abstract}
		Capsule networks (CapsNets) are capable of modeling visual hierarchical relationships, which is achieved by the ``routing-by-agreement'' mechanism.
		This paper proposes a pairwise agreement mechanism to build capsules, inspired by the feature interactions of factorization machines (FMs). 
		The proposed method has a much lower computation complexity.
		We further proposed a new CapsNet architecture that combines the strengths of residual networks in representing low-level visual features and CapsNets in modeling the relationships of parts to wholes.
		We conduct comprehensive experiments to compare the routing algorithms, including dynamic routing, EM routing, and our proposed FM agreement, based on both architectures of original CapsNet and our proposed one, and the results show that our method achieves both excellent performance and efficiency under a variety of situations. 
	\end{abstract}

	\section{INTRODUCTION}
	Encoding entities by vectors is widely practiced in the deep  models, especially the domains like Natural Language Processing (NLP) and recommendation system. 
	However, this idea is not often used in computer vision tasks such as image classification, and the CapsNets are representative in terms of ``vector-wise'', which encode objects by collections of neurons called capsules.
	It is introduced to address the problem of information loss, such as position, size, rotation, scale, which is caused by pooling layers~\cite{auto-encoders}. 
	A capsule represents an object and is often composed of a pose vector/matrix and an activation, where the pose vector/matrix encodes the instantiation parameters of this object. This design aims to disentangle the pose from the evidence of its existence. 
	When the viewing condition changes, the instantiation parameters change, but the capsule still stays active. Such a property is called equivariance and invariance~\cite{auto-encoders,dynamicrouting}, and can be used to build visual hierarchical relationships between capsules of different layers with a characteristic of assigning parts to wholes.
	
	\begin{figure}[t]
		%	\begin{center}    %居中
		\includegraphics[width=0.9\linewidth]{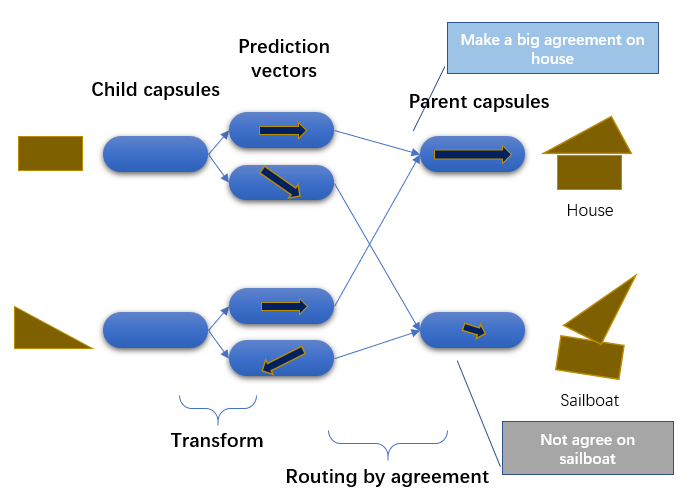}
		%	\end{center}
		\caption{Illustration of the mechanism of routing-by-agreement. We show how the child capsules agree on ``House'' instead of ``Sailboat''. } 
		\label{fig:routing-by-agreement}  %图片引用标记
	\end{figure}
	
	\subsection{Routing-by-Agreement Mechanism}
	Sabour \etal~\cite{dynamicrouting} introduce dynamic routing to achieve visual hierarchical relationships in CapsNets. 
	It iteratively routes information from lower-level capsules to upper-level ones by a mechanism called ``routing-by-agreement'', and this is the critical idea of dynamic routing in CapsNets.
	Hinton \etal~\cite{auto-encoders} take an example to explain the ``agreement'':
	\begin{quote}
		If, for example, A represents a mouth and B represents a nose, they can each make a prediction for the pose of the face. If these predictions agree, the mouth and nose must be in the right spatial relationship to form a face. 
	\end{quote} 
	
	Figure \ref{fig:routing-by-agreement} shows an intuitive description to illustrate the mechanism of routing-by-agreement,  where each capsule is a vector, and the inner product between them denotes how much they agree with each other. The child capsules make an agreement that ``House'' is the right prediction instead of ``Sailboat''.
	Here comes the core problem of how to compute the ``agreement'' from these prediction vectors (produced by the child capsules in the lower-layer).  
	The original paper~\cite{dynamicrouting} implements this by iteratively computing the vectors' inner product (agreement) between prediction vectors and their corresponding parent capsules in the upper-layer.
	The more the prediction vectors agree (in terms of vectors' inner product) with a parent capsule, the parent capsule gets a larger activation. 
	The vector length of this parent capsule encodes how strong the child capsules agree with it. 
	However, this implementation has limited performance and an expensive cost both in terms of memory and computation~\cite{segcaps,Encapsulation,deepcaps}, especially with multiple routing iterations.
	The CapsNet proposed by Sabour \etal~\cite{dynamicrouting} works well with the MNIST~\cite{lecun1998mnist} dataset but is not on par with traditional CNNs on more complex datasets such as CIFAR10~\cite{CIFAR10}.
	The convolutional CapsNet with EM routing~\cite{EMRouting} achieves an impressive performance on smallNORB~\cite{lecun2004learning} dataset, which demonstrates the capability of modeling viewpoints invariance.
	However, there are not enough experiments on other kinds of datasets. Moreover, it suffers from  a much higher cost of memory and computation than the CapsNet in~\cite{dynamicrouting}.
	These problems largely limit the practice of CapsNets to be applied to more challenge tasks.

	\subsection{``Agreement'' in FMs}
	The prediction of click-through rate (CTR) is to estimate the probability a user clicks on an item in recommendation systems, and it is critical to learn feature interactions behind user click behaviors, such as app category and time-stamp. 
	Rendle and Steffen~\cite{fm} introduced Factorization Machines (FMs) to capture feature interactions automatically and showed promising results on input features which are sparse and of enormous dimension. 
	FMs model feature interactions as the inner product of latent vectors, which is at ``vector-wise'' level instead of ``bit-wise'', this is somewhat corresponding to the capsules in CapsNets. 
	%And more importantly, previous work \cite{fm,collaborative_filtering} has shown that local %dependencies between features from different fields can be effectively explored by feature %vector ``product'' operations instead of ``add'' operations. 
	FMs usually consider only order-2 feature interactions in practice.
	%instead of high-order feature interactions due to high complexity. 
	Each feature is embedded into a latent factor vector, and the pairwise feature interactions are modeled as the inner products of latent vectors.
	%: $\sum_{i=1}^{n} \sum_{j=i+1}^{n}\left\langle\mathbf{v}_{i}, \mathbf{v}_{j}\right\rangle$. 
	This is further reformulated by Rendle and Steffen~\cite{fm} to make its computation more efficient.
	%in linear time $O(k n)$.
	
	Inspired by FMs, we find that the way it models the pairwise feature interactions can also be used to model the agreements (in terms of pairwise vector inner product) of prediction vectors in CapsNets. 
	The sum of these agreements (each pair of prediction vectors) measures how much the prediction vectors agree with each other. The more they agree with each other, the more the corresponding parent capsule gets activated. 
	An FM algorithm gets much more efficient since it
	does not need to iteratively compute the parent capsule and its agreements with the prediction vectors. 
	Our contributions include:
	\begin{itemize}
		\item We propose an efficient routing-by-agreement algorithm called FM agreement, which is inspired by the idea of modeling feature interactions in FMs, which outperforms the original dynamic routing and EM routing with a much lower computation complexity.
		\item We further propose a novel CapsNet architecture that contains residual network blocks and capsule layers, and this architecture turns out to successfully combines both strengths of CapsNets and ResNets.
		\item We conduct several experiments on benchmark datasets. The results demonstrate that our approaches achieve better performance and efficiency than baselines while retaining the properties of CapsNets.
	\end{itemize}

	\section{RELATED WORK}
	The concept of capsules was firstly introduced by Hinton \etal~\cite{auto-encoders} to address the representational limitations. 
	Sabour \etal~\cite{dynamicrouting} proposed CapsNet with dynamic routing algorithm, which achieved a state-of-the-art result on MNIST dataset. 
	Hinton \etal~\cite{EMRouting} proposed a new iterative routing procedure based on the EM algorithm, which achieved an impressive result on smallNORB dataset. 
	After that, the idea of the capsule was applied to many specific tasks to improve performance~\cite{VideoCapsuleNet,segcaps,Capsule-text}. 
	Many efforts~\cite{Spectral-capsule,Encapsulation,DCNet,An-optimization-view} have been made to seek better capsule architectures.
	Shahroudnejad \etal~\cite{Relevance-path-by-agreement} focused on investigating the potential explainability properties of CapsNets.
	Lenssen \etal~\cite{Group-Equivariant-Capsule} proposed ``Group-Equivariant-Capsule'' to introduce guaranteed equivariance and invariance properties to the CapsNets.
	It interestingly applied the idea of Lie groups, but with limitations that only a few realizations are applicable in a deep neural network architecture.
	Recently, Rajasegaran \etal~\cite{deepcaps} explored CapsNets by ``going deeper'' and proposed DeepCaps, which aimed at improving the performance of the CapsNets for more complex image datasets. It achieved some improvement on the CIFAR10 dataset compared to the work of Sabour \etal~\cite{dynamicrouting}, but which was still very limited when compared to the traditional CNNs such as ResNets~\cite{resnet}.
	
	Rendle \etal~\cite{fm} introduced Factorization Machines (FMs) to model feature interactions automatically, it achieved great success in the problem of click-through rate (CTR).
	Many new architectures related to the idea of FMs have been proposed in recent years, the representative works include FNN~\cite{FNN}, PNN~\cite{PNN}, DeepCross~\cite{deepcross}, NFM~\cite{NFM}, DCN~\cite{dcn}, DeepFM~\cite{deepfm}, and xDeepFM~\cite{xdeepfm}.
	
	Different from the above works, we are the first to introduce the idea of pairwise feature interactions in FMs as a routing-by-agreement mechanism to improve the CapsNets, which aims to provide better performance and efficiency for CapsNets.
	Our work also explores the capsule architectures, which is related to the work in~\cite{deepcaps} that explores how to construct deeper CapsNets. However, our work has some differences to the work in~\cite{deepcaps}: 1) Our work focuses on combining the advantages of  CNNs and CapsNets,  to achieve better performance and retain the properties of CapsNets, while the work in~\cite{deepcaps} only construct CapsNets by stack multiple capsule layers; 2) Our work further designs a new routing-by-agreement algorithm. 
	%We also did exploration on the capsule architectures, and agree with~\cite{deepcaps} that going deeper with CapsNet is a step in the right direction to further enhance the performance, but not just  to make it deeper, we believe the CNN backbones are also a key component, by which we can combine both strengths of CapsNet and traditional CNNs.
	
	\section{METHOD}
	For the CapsNet in~\cite{dynamicrouting}, each child capsule is denoted by a vector $\mathbf{u}_i$, and it first computes a ``prediction vector'' $\hat{\mathbf{u}}_{j|i}$ for each possible parent capsule by multiplying a weight matrix $\mathbf{W}_{ij}$:
	\begin{align}
	\hat{\mathbf{u}}_{j | i}=\mathbf{W}_{i j} \mathbf{u}_{i}.
	\end{align}
	Given these $\hat{\mathbf{u}}_{j|i}$, an agreement mechanism is performed to compute the parent capsules $\hat{\mathbf{u}}_{j}$  in the upper-layer, which is the crucial point we investigate in this paper. 
	One way to achieve agreement is dynamic routing  \cite{dynamicrouting}, where the  $\hat{\mathbf{u}}_{j}$  is computed iteratively from a weighted average of prediction vectors $\hat{\mathbf{u}}_{j|i}$, and the agreements are computed by vector inner product between $\hat{\mathbf{u}}_{j}$ and each $\hat{\mathbf{u}}_{j|i}$:
	\begin{align}\label{eqn:agreement}
	a_{ij}=\left\langle\hat{\mathbf{u}}_{j|i},\hat{\mathbf{u}}_{j}\right\rangle.
	\end{align}
	These agreements are used to compute the coupling coefficients $c_{ij}$
	%(which denote the weights used for the next iteration of weighted average operation) 
	for $\hat{\mathbf{u}}_{j|i}$ in the next iteration. 
	Here the coupling coefficients  $c_{ij}$ measure how much the $\hat{\mathbf{u}}_{j}$ and $\hat{\mathbf{u}}_{j|i}$ agree, or how much a child capsule $\hat{\mathbf{u}}_i$ coupled to a parent one $\hat{\mathbf{u}}_j$.
	
	%The original dynamic routing iteratively computing the coupling coefficients $c_{ij}$ from vector inner product(agreement): $a_{ij}=\hat{\mathbf{u}}_{j}\dot{\mathbf{v}_j}$ 
	%logits $b_{ij}$ from vector inner product between $\hat{\mathbf{u}}_{j|i}$ and $\mathbf{v}_j$, $c_{ij}$ is the coupling coefficients computed from $b_{ij}$ by $\mathit{softmax}$ to make sure they sum to 1 along the dimension$j$, it is clear that $c_{ij} \propto b_{ij}$, and from the line 4 we can get:
	%\begin{align}
	%\begin{aligned}
	%	\left\|\mathbf{s}_j\right\|&=\left\|\sum_{i} c_{i j} \hat{\mathbf{u}}_{j | i}\right\| \\
	%	&=
	%\end{aligned}
	%\end{align}
	When certain parent capsule $\hat{\mathbf{u}}_{j}$  couples much to multiple child capsules $\mathbf{u}_i$, this parent capsule becomes activated. The main problem of this dynamic routing algorithm is its expensive cost, both in terms of memory and computation \cite{segcaps,Encapsulation,deepcaps}. Moreover,  it is of limited performance compared with traditional CNNs like Residual networks~\cite{resnet}.

	\subsection{Agreement Mechanism in FMs}
	Given feature vectors $\{\mathbf{v}_i\}_{i=1}^n$ with a size of $n$, the FMs~\cite{fm} model the pairwise feature interactions as: 
	\begin{align}\label{eqn:fm1}
	y :=\sum_{i=1}^{n} \sum_{j=i+1}^{n}\left\langle\mathbf{v}_{i}, \mathbf{v}_{j}\right\rangle x_{i} x_{j},
	\end{align}
	where $\langle\cdot, \cdot\rangle$ is the inner product of two vectors of size $k$: $\left\langle\mathbf{v}_{i}, \mathbf{v}_{j}\right\rangle :=\sum_{f=1}^{k} v_{i, f} \cdot v_{j, f}$. Here
	$\mathbf{v}_{i}$ describes the $i$-th variable with $k$ factors. $k \in \mathbb{N}_{0}^{+}$ is a hyperparameter that defines the dimensionality of the factorization, which is also corresponding to the number of neurons in a capsule of this paper. $\left\langle\mathbf{v}_{i}, \mathbf{v}_{j}\right\rangle$ models the interaction between the $i$-th and $j$-th variable.
	$x_i$ and $x_j$ are binary, which denote whether the corresponding $\mathbf{v}_{i}$, $\mathbf{v}_{j}$ are activated. So $y$ is the sum of the pairwise interactions of the activated vectors.
	The computational complexity  of Formula \ref{eqn:fm1} is $O(k n^2)$ since all the pairwise interactions have to be computed.
	For relieving the computation problem, the pairwise interactions can be reformulated as~\cite{fm}: 
	\begin{align}
	\label{eqn:simplified fm}
	\begin{aligned} & \sum_{i=1}^{n} \sum_{j=i+1}^{n}\left\langle\mathbf{v}_{i}, \mathbf{v}_{j}\right\rangle x_{i} x_{j} \\=& \frac{1}{2} \sum_{f=1}^{k}\left(\left(\sum_{i=1}^{n} v_{i, f} x_{i}\right)^{2}-\sum_{i=1}^{n} v_{i, f}^{2} x_{i}^{2}\right) \end{aligned}
	\end{align}
	This equation has a computation complexity of  $O(k n)$, which is linear in both $k$ and $n$. 
	
	\subsection{Pairwise Agreement in Capsules}
	Given a bunch of prediction vectors $\left[\hat{\mathbf{u}}_{j|0}, \dots, \hat{\mathbf{u}}_{j|n},\right]$,
	we propose to model the ``agreement'' between capsules following the idea of pairwise interactions as FMs do. 
	The original FM algorithm computes the pairwise interactions by inner product as: $	\hat{a}_{j|i_1,i_2}=\left\langle\hat{\mathbf{u}}_{j|i_1},\hat{\mathbf{u}}_{j|i_2}\right\rangle$, which only models the magnitude of the agreements. 
	Here, we propose to use the element-wise product as: $\hat{\mathbf{a}}_{j|i_1,i_2}=\hat{\mathbf{u}}_{j|i_1} \odot \hat{\mathbf{u}}_{j|i_2}$. 
	Such a method not only can derive the magnitude of the agreements by summing over each element of $\hat{\mathbf{a}}_{j|i_1,i_2}$, but also can represent the orientation.
	Specifically, we first compute the pairwise interactions of capsules as:
	\begin{align}
	\begin{aligned}
	\label{eqn:ss_j}
	\hat{\mathbf{s}}_j & = \sum_{i_1=1}^{n} \sum_{i_2=i_1+1}^{n} \hat{\mathbf{u}}_{j|i_1} \odot \hat{\mathbf{u}}_{j|i_2}\\
	& =\frac{1}{2} \left(\sum_{i=1}^{n}\hat{\mathbf{u}}_{j|i}\odot \sum_{i=1}^{n}\hat{\mathbf{u}}_{j|i}-\sum_{i=1}^{n}\hat{\mathbf{u}}_{j|i}\odot\hat{\mathbf{u}}_{j|i}\right) \end{aligned}
	\end{align}
	where $\hat{\mathbf{u}}_{j|i}=\left[\hat{u}_{j|i,1},\dots \hat{u}_{j|i,k}\right]$, $\hat{\mathbf{s}}_j=\left[\hat{s}_{j,1},\dots \hat{s}_{j,k}\right]$, and $n$ denotes the number of prediction vectors along the dimension $i$.
	%, $k$ denotes the number of neurons in capsule $\hat{\mathbf{u}}_{j}$. %operation $\odot$ is the element-wise product between vectors.
	Then the activation of the output capsule $\hat{\mathbf{u}}_{j}$ (the ``agreement'' of capsules $\mathbf{u}_i$ on $\hat{\mathbf{u}}_{j}$) can be formulated as:
	\begin{align}\label{eqn:a_j}
	\hat{a}_j=\sum_{f=1}^{k} \hat{s}_{j,f}
	\end{align}
	We define the pose vector as $\hat{\mathbf{p}}_j=\frac{\hat{\mathbf{s}}_j}{\left\|\hat{\mathbf{s}}_j\right\|}$. The direction of $\hat{\mathbf{p}}_j$ encodes the properties of an entity such as position, size, rotation, scale, $etc.$ 
	
	We derive the partial of $\hat{a}_j$ with respect to $\hat{u}_{j|i, f}$:
	\begin{align}
	\label{eqn:jaco}
	\begin{aligned} \frac{\partial \hat{a}_j}{\partial \hat{u}_{j|i, f}} &= \frac{\partial \hat{s}_{j,f}}{\partial \hat{u}_{j|i, f}} \\
	&=\frac{1}{2} \left(\frac{\partial\left(\sum_{i=1}^{n} \hat{u}_{j|i, f}\right)^{2}}{\partial \hat{u}_{j|i, f}}-\frac{\partial \sum_{i=1}^{n} \hat{u}_{j|i, f}^{2}}{\partial \hat{u}_{j|i, f}}\right) \\
	&=\sum_{i=1}^{n} \hat{u}_{j|i, f}-\hat{u}_{j|i, f} \end{aligned}
	\end{align}
	
	In practice, we observe that the routing can result in activation/gradient explosion and the potential numeric problem, which is mainly caused by the sum operation shown in Eqn.\ref{eqn:ss_j} and~\ref{eqn:a_j}. 
	To avoid it, we argue the key is to ensure the singular value of the Jacobian of $\frac{\partial \hat{a}_j}{\partial \hat{\mathbf{u}}_{j|i}}$ to be near one.
	From Eqn.~\ref{eqn:jaco}, we observe that the value of $\sum_{i=1}^{n} \hat{u}_{j|i, f}$ makes the most contribution to the gradient. We thus scale the $\hat{\mathbf{u}}_{j|i}$ by dividing $\sqrt{n}$, which results in:
	\begin{align}\label{eqn:s_j}
	\hat{\mathbf{s}}_j & = \frac{1}{2n} \left(\sum_{i=1}^{n}\hat{\mathbf{u}}_{j|i}\odot \sum_{i=1}^{n}\hat{\mathbf{u}}_{j|i}-\sum_{i=1}^{n}\hat{\mathbf{u}}_{j|i}\odot\hat{\mathbf{u}}_{j|i}\right)
	\end{align}
	We find Eqn.~\ref{eqn:s_j} well remedies the gradient explosion problem. 
	Furthermore, we apply L2 normalization to the $\hat{\mathbf{u}}_{j|i}$ to make its length be one before agreement routing operation.
	This process guarantees the agreement value is computed by the direction of vectors and makes the training more stable according to our experiments.
	We conclude the above process in Algorithm \ref{alg:routing-by-fm}, referred to as {\bf FM Agreement}.
	
	\begin{algorithm}[tb]
		\setstretch{1.4}
		\caption{FM Agreement}
		\label{alg:routing-by-fm}	
		\begin{algorithmic}[1] %[1] enables line numbers
			\REQUIRE prediction vectors $\hat{\mathbf{U}}_j=\left(\hat{\mathbf{u}}_{j|1},\dots,\hat{\mathbf{u}}_{j|n}\right)$ %$\hat{\mathbf{u}}_{j|i}$
			\ENSURE $\hat{\mathbf{p}}_j$, $\hat{a}_j$ 
			
			\STATE $\hat{\mathbf{u}}_{j|i} \gets \textit{L2Normalize}(\hat{\mathbf{u}}_{j|i})$ \hfill $\forall i$
			\STATE $\hat{\mathbf{s}}_j\gets \frac{1}{2n} \left(\sum_{i=1}^{n}\hat{\mathbf{u}}_{j|i}\odot \sum_{i=1}^{n}\hat{\mathbf{u}}_{j|i}-\sum_{i=1}^{n}\hat{\mathbf{u}}_{j|i}\odot\hat{\mathbf{u}}_{j|i}\right)$
			\STATE $\hat{\mathbf{p}}_j\gets \frac{\hat{\mathbf{s}}_j}{\left\|\hat{\mathbf{s}}_j\right\|}$
			\STATE $\hat{a}_j\gets  \sum_{f=1}^{k}\hat{s}_{j,f}$
		\end{algorithmic}
	\end{algorithm}
	
	We use $\hat{a}_j$ as the prediction of each category for the image classification tasks, where $\hat{a}_j$ indicates how much this category is activated. In the subsequent experiments, we use the softmax cross entropy as the loss function just like the models in~\cite{resnet,densenet,vgg,googlenet}, for the tasks which predict only one right category (\eg, the classification task shown in Section \ref{sec:ex2}, \ref{sec:ex3-1}, \ref{sec:ex4}). 
	Considering the classification tasks where the number of categories in the image is uncertain (\eg, the experiments in Section \ref{sec:ex3-2}), we use the ``Margin Loss'' ~\cite{dynamicrouting}:
	\begin{align}\label{eqn:margin loss}
	\begin{aligned}L_{j}=&T_{j} \max \left(0, m^{+}-\hat{a}_j\right)^{2}+\\&
	\lambda\left(1-T_{j}\right) \max \left(0,\hat{a}_j-m^{-}\right)^{2}, \end{aligned}
	\end{align}
	%by which we can set a threshold to indicate whether the category represented by the capsule exists. 
	where $T_j=1$ if an object of class $j$ is present. We set $\lambda=0.5, m^{+}=0.9, m^{-}=0.1$ by default. The $\lambda$ is used to down-weight the loss for absent classes.
	
	\subsection{Architecture of One Capsule Layer} 
	To reduce the parameters, we compute the prediction vectors $\hat{\mathbf{u}}_{j|i}$ as the work in~\cite{EMRouting} does.
	The capsule described in this paper is represented as a vector of $k$ dimensions. 
	For adapting the method in computing the prediction, we need to reshape it into a $k^{'}\times k^{'}$ matrix where $k^{'}$ should be equal to $\sqrt{k}$, and ensure the trainable weight matrix $\mathbf{W}_{ij}$ between capsule $i$ in layer $L$ and capsule $j$ in layer $L+1$ being also a $k^{'}\times k^{'}$ matrix~\cite{EMRouting}.
	By doing this,  we get the ``prediction matrix'' by multiplying the reshaped $\mathbf{u}_i$ and $\mathbf{W}_{ij}$, then reshape it into a $k$ dimensions vector $\hat{\mathbf{u}}_{j|i}$ which is used as the input of the routing algorithm.
	One advantage of this operation is that we reduce the parameters from $k\times k$ to $\sqrt{k}\times \sqrt{k}$,
	% by (\ie, to transform a $k$ dimensions child capsule $\mathbf{u}_i$ in layer $L$ into a $k$ dimensions prediction vector $\hat{\mathbf{u}}_{j|i}$, the parameters are reduced , 
	and there is no performance degeneration, according to our experiments.
	We set $k$ as 16 in the experiments by default.
	Before the routing operation, we insert a batch normalization layer~\cite{ioffe2015batch} to normalize the prediction vectors, which  stabilize the training process. The transformation process can be concluded in Algorithm \ref{alg:transformation}.
	\begin{algorithm}[tb]
		\setstretch{1.2}
		\caption{Matrix Transformation}
		\label{alg:transformation}	
		
		\begin{algorithmic}[1] %[1] enables line numbers
			\REQUIRE $\mathbf{u}_{i}$, $\mathbf{W}_{ij}$
			\ENSURE $\hat{\mathbf{u}}_{j|i}$
			
			\STATE $\mathbf{u}_{i}\gets \textit{Reshape}(\mathbf{u}_{i})$ \hfill $1\times k \rightarrow \sqrt{k}\times \sqrt{k}$
			\STATE $\hat{\mathbf{u}}_{j|i}\gets \mathbf{u}_{i}\times \mathbf{W}_{ij}$ \hfill Matrix multiplication
			\STATE $\hat{\mathbf{u}}_{j|i}\gets \textit{Reshape}(\hat{\mathbf{u}}_{j|i})$ \hfill $\sqrt{k}\times \sqrt{k}\rightarrow 1\times k$
			\STATE $\hat{\mathbf{u}}_{j|i}\gets \textit{BatchNormalization}(\hat{\mathbf{u}}_{j|i})$
		\end{algorithmic}
	\end{algorithm}

\begin{figure}[bt]
		\begin{center} 
			\includegraphics[width=0.9\linewidth]{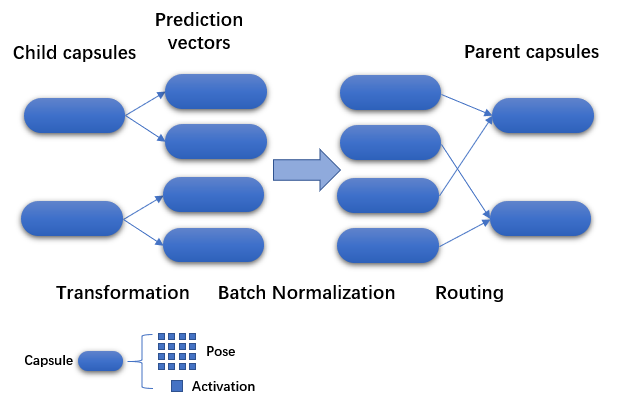}
		\end{center}
		\caption{Illustration of one capsule layer. The prediction vectors of input child capsules, corresponding to each parent capsule,  are firstly  computed,  and then  transformed into the output parent capsules by a routing procedure.} 
		\label{fig:capsule layer}  %图片引用标记
	\end{figure}

	Figure \ref{fig:capsule layer} presents the architecture of a capsule layer. 
	A capsule contains a pose vector/matrix and an activation, and there are different ways to implement it~\cite{EMRouting,dynamicrouting}. 
	In the implementation of~\cite{dynamicrouting}, the activation is already encoded in the pose vector (because it's defined as the length of the pose vector), and the dynamic routing algorithm only takes the pose vector as its inputs.
	While in the implementation of~\cite{EMRouting}, they are separated, and the EM routing algorithm takes both pose matrix and activation as its inputs (in which the activation is used in the ``m-step''). 
	To adapt our FM agreement to the multiple capsule layers, we make the $\hat{\mathbf{s}}_j$ in Formula \ref{eqn:s_j} as output capsules of the current layer, which are also the input capsules of the next layer.

	\subsection{Combining with Traditional CNNs}\label{sec:combining}
	The CapsNets are designed to be a new paradigm to address the problem, such as information loss in traditional neural network architectures.
	%or the higher levels of a visual system, it is used to solve the problem of assigning parts to wholes.
	According to the implementations of CapsNets in~\cite{EMRouting,dynamicrouting}, the capsule layers are built upon a few CNN layers which are used to learn lower level features, and these lower level features are further used by capsule layers to learn parse trees and relationships of parts to wholes~\cite{dynamicrouting}. 
	Therefore, the quality of these lower level features does affect the performance of upper capsule layers. 
	Besides, capsule layers are computation expensive, especially applied in lower layers with a large spatial dimension~\cite{deepcaps}.
	Rajasegaran \etal reduce the computation complexity in their deep capsule model by reducing the number of routing iterations in the lower layers.
	At the same time, the performance is not affected as the features in the lower layers need not be complex in nature.
	We believe these lower layers can be replaced by more efficient components.
	Traditional CNN architectures like VGG~\cite{vgg}, ResNet~\cite{resnet}, GoogLeNet~\cite{googlenet}, DenseNet~\cite{densenet}, and their many variants have achieved huge success in many computer vision tasks, by learning better visual representation. 
	One straightforward idea is to combine the strengths of both traditional CNNs and CapsNets.
	
\begin{figure}[tb]
		\begin{center}    %居中
			\includegraphics[width=0.9\linewidth]{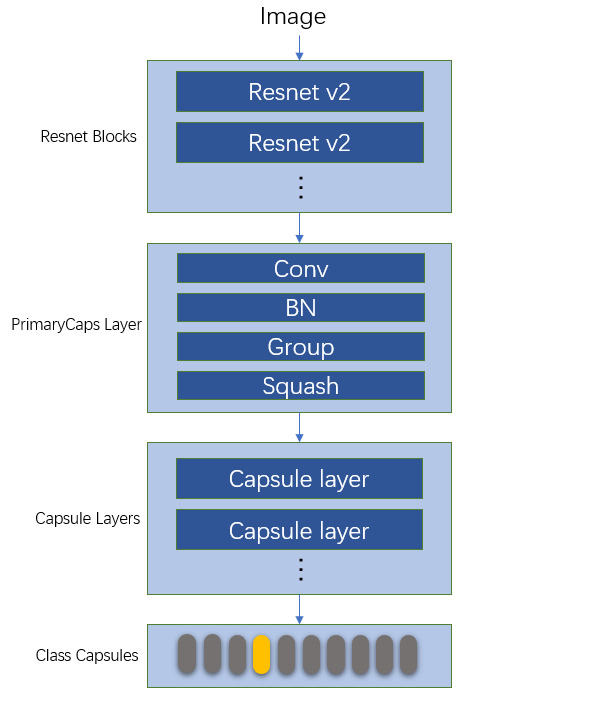}
		\end{center}
		\caption{The architecture of the proposed CapsNet. The input image is first fed into the ResNet blocks to produce feature maps, and then transformed into capsules by ``PrimaryCaps'' layer. These input capsules are used by the following capsule layers to generate output capsules.} 
		\label{fig:complete arch}  %图片引用标记
	\end{figure}

	We design the architecture with both ResNet blocks and capsule layers, and such an architecture can be used to investigate the performance of networks and simultaneously explore the properties of the capsule.
	We apply several ResNet-v2~\cite{he2016identity} blocks as the backbone and use a ``PrimaryCaps layer'' to connect the backbone and its above capsule layers.
	Inside this PrimaryCaps layer, the tensor provided by the backbone is first to be downsampled by a convolutional layer, then normalized by a batch normalization layer. 
	This tensor is further grouped into capsules by reshaping (\eg, a tensor with shape $H\times W\times C$ is reshaped into $H\times W\times C^{'}\times K$, where $H, W, C$ indicates height, width, channel respectively, and $K$ is the neuron number in a capsule), and activated by a non-linear ``squashing''~\cite{dynamicrouting} function. 
	We describe the complete architecture in Figure \ref{fig:complete arch}.

	\begin{figure*}[tbh]
		\centering
		\subfigure[cropped-SVHN]{
			\begin{minipage}[c]{.32\linewidth}
				\centering
				\includegraphics[width=6cm]{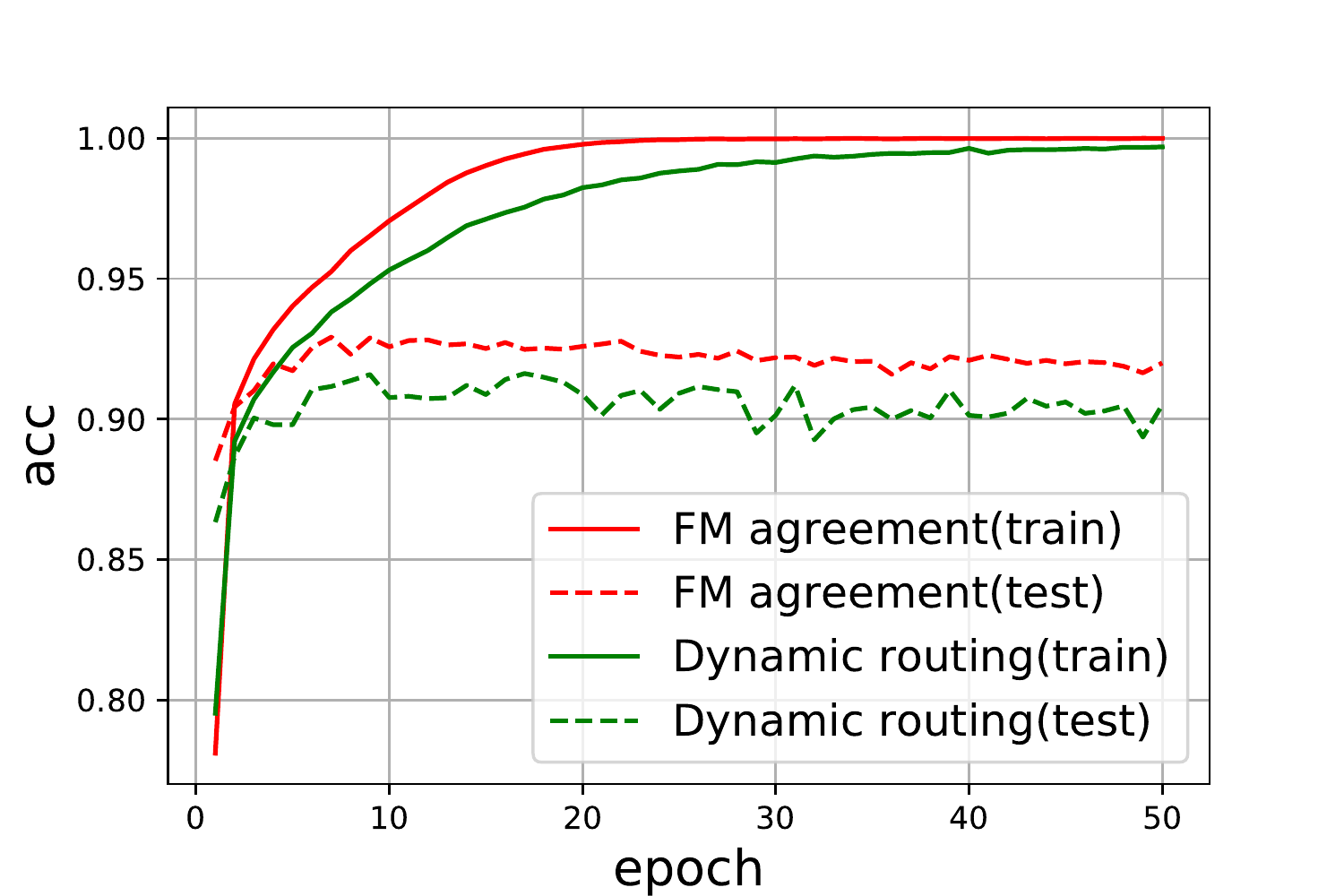}
				\vspace{0.03in}
			\end{minipage}
		}
		\subfigure[Fashion-MNIST]{
			\begin{minipage}[c]{.32\linewidth}
				\centering
				\includegraphics[width=6cm]{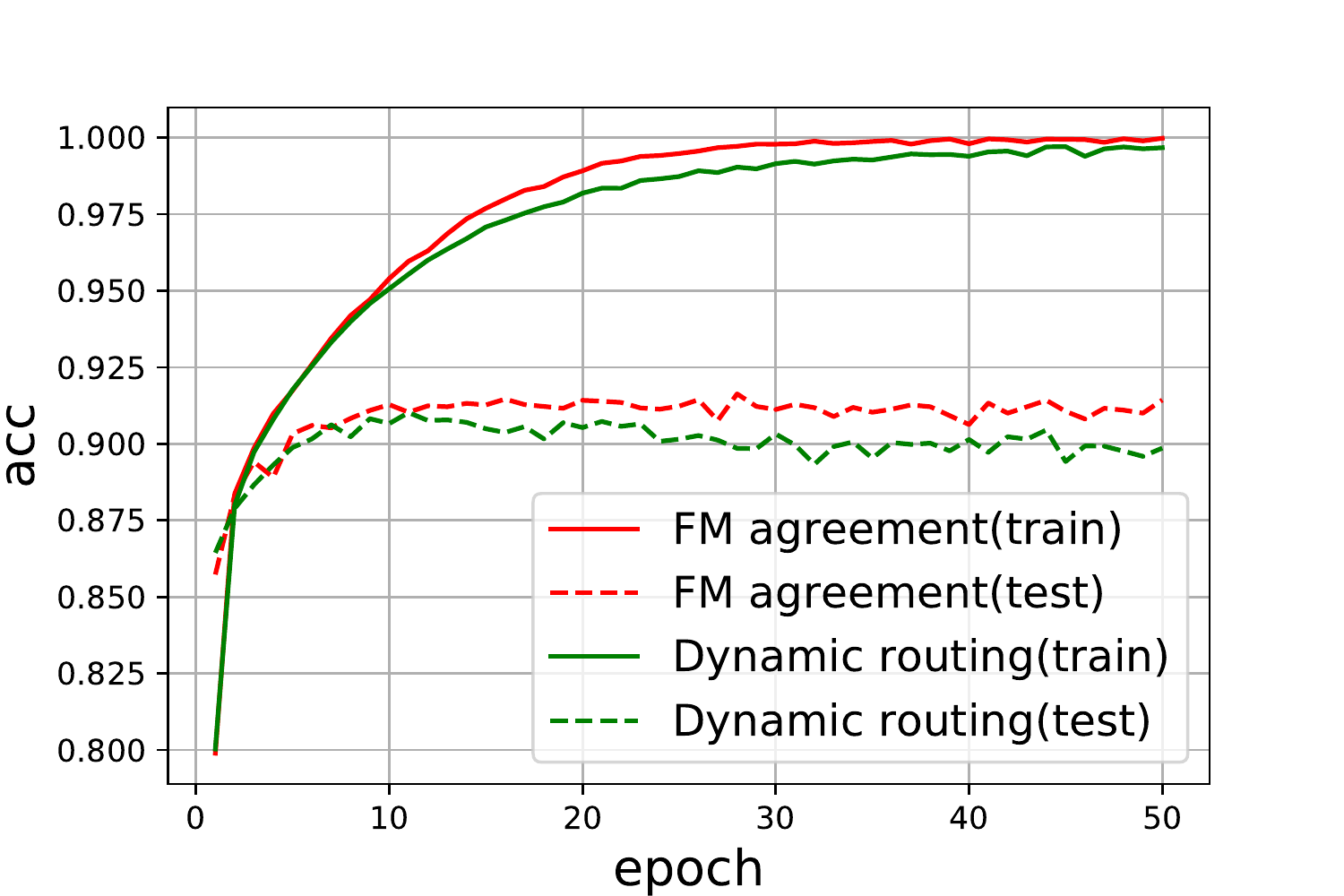}
				\vspace{0.03in}
			\end{minipage}
		}
		\subfigure[CIFAR10]{
			\begin{minipage}[c]{.32\linewidth}
				\centering
				\includegraphics[width=6cm]{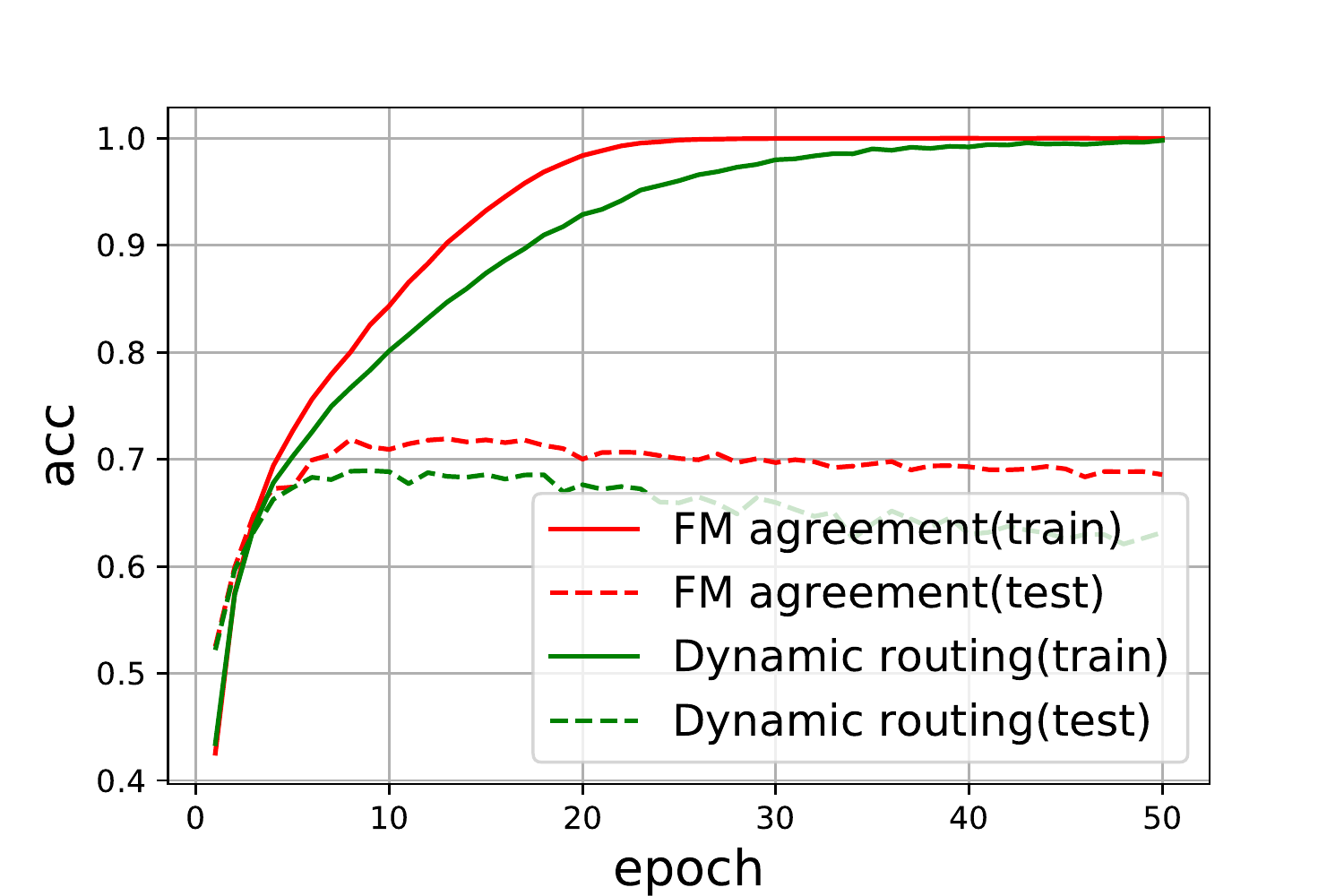}
				\vspace{0.03in}
			\end{minipage}
		}
		\vspace{-0.1in}
		\caption{Comparison of FM agreement and dynamic routing without data augmentation.}
		\label{fig:ori1}
		\vspace{-0.2in}
	\end{figure*}
	
	\begin{figure*}[htb]
		\centering
		\subfigure[cropped-SVHN+random-crop]{
			\begin{minipage}[c]{.32\linewidth}
				\centering
				\includegraphics[width=6cm]{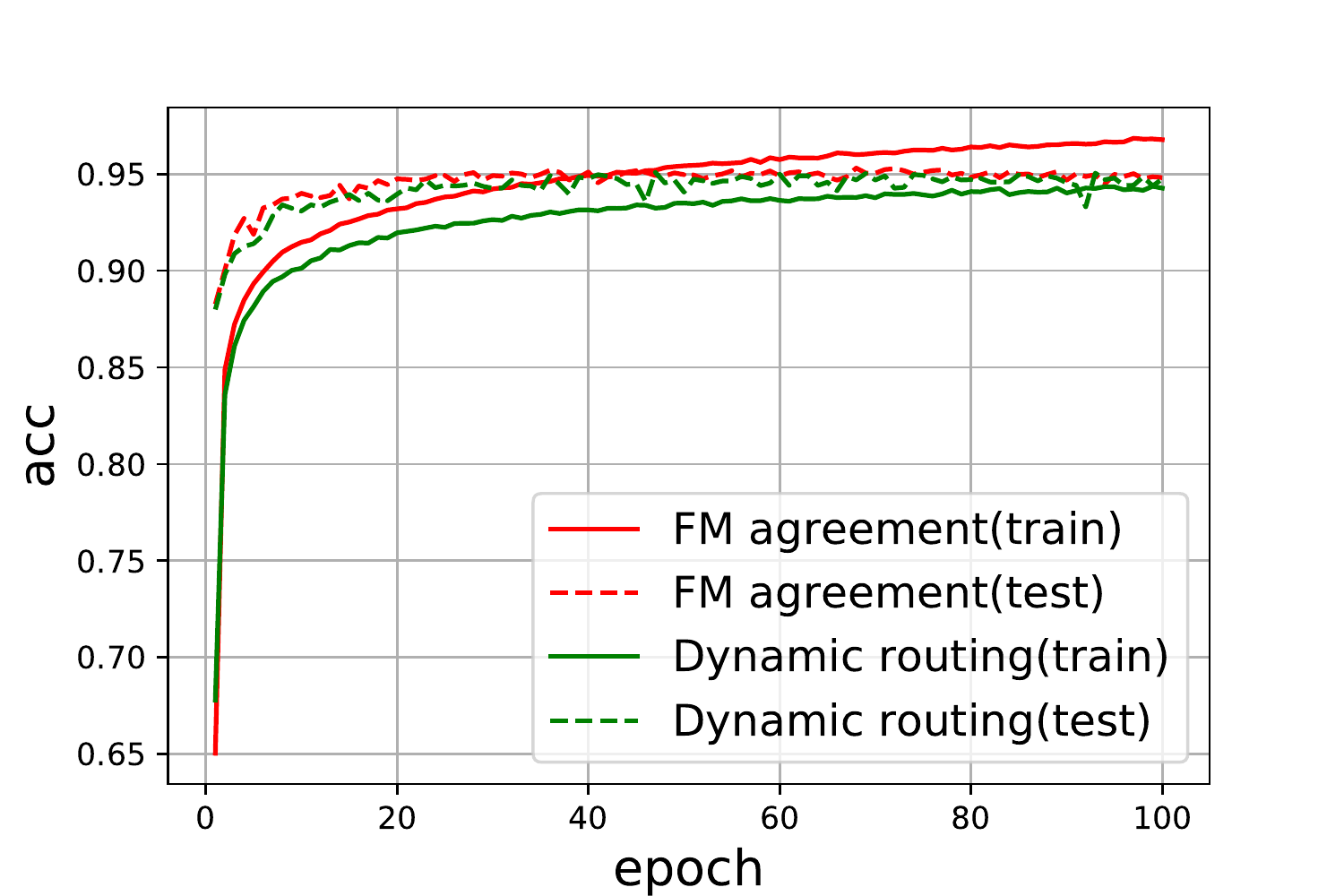}
				\vspace{0.03in}
			\end{minipage}
		}
		\subfigure[Fashion-MNIST+random-crop]{
			\begin{minipage}[c]{.32\linewidth}
				\centering
				\includegraphics[width=6cm]{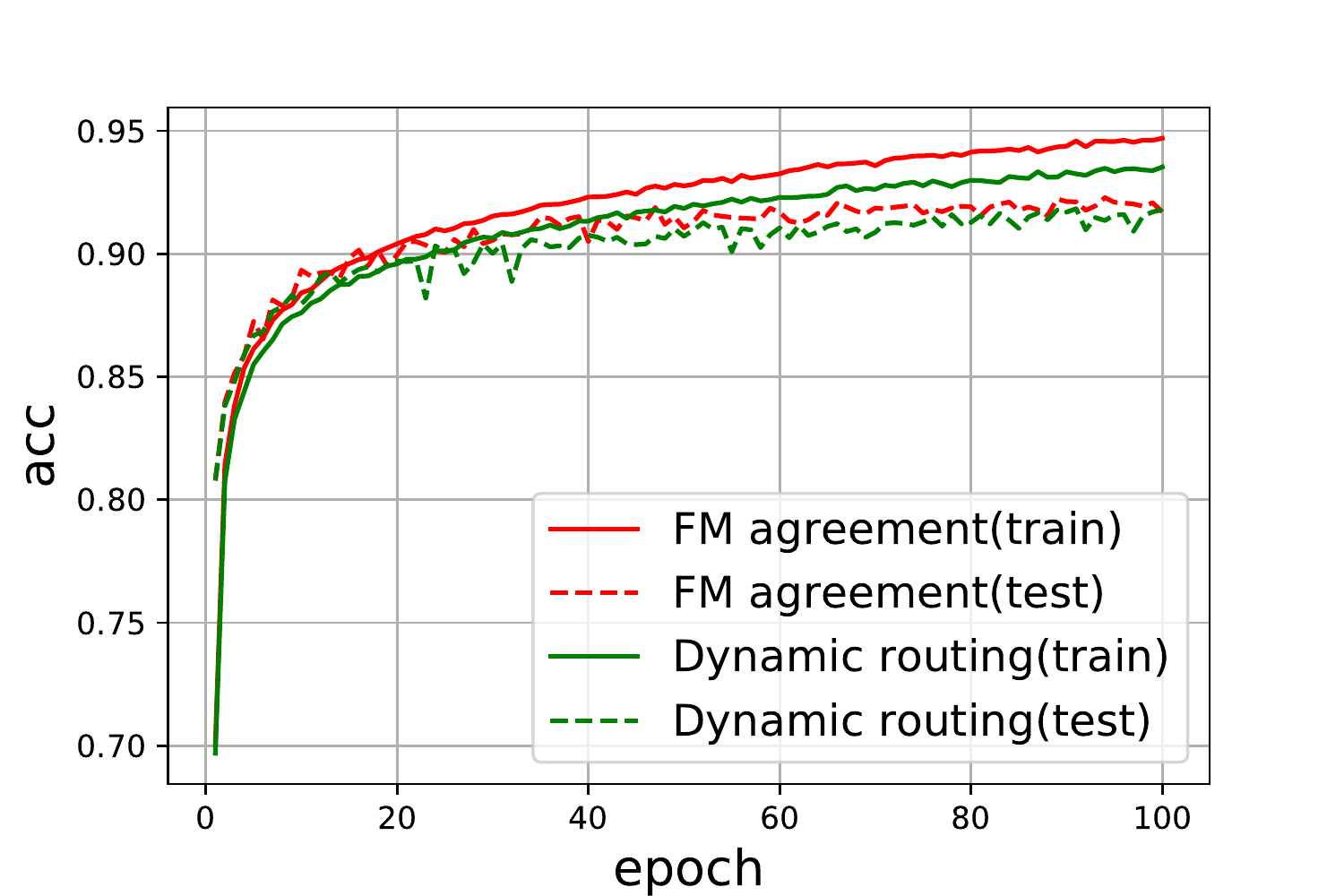}
				\vspace{0.03in}
			\end{minipage}
		}
		\subfigure[CIFAR10+random-crop+random-flip]{
			\begin{minipage}[c]{.32\linewidth}
				\centering
				\includegraphics[width=6cm]{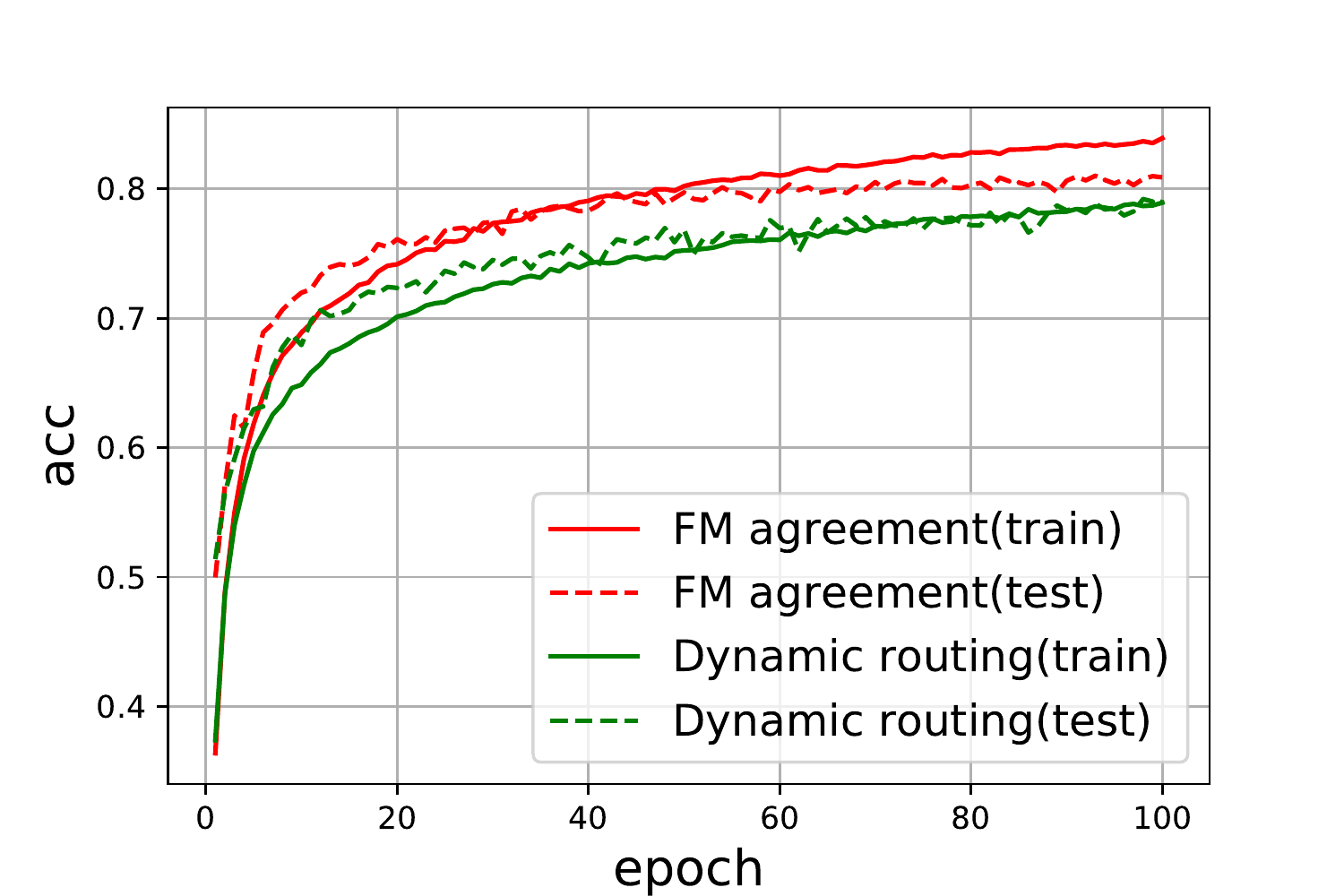}
				\vspace{0.03in}
			\end{minipage}
		}
		\vspace{-0.1in}
		\caption{Comparison of FM agreement and dynamic routing with data augmentation.}
		\label{fig:ori2}
	\end{figure*}

	\section{EXPERIMENTS}
	In this section, we conduct experiments to validate the effectiveness of our method in image classification and maintaining the equivariance and invariance.
	
	We implement all the models by Tensorflow,\footnote{The code is available at \url{https://github.com/bhneo/FMRouting}.} and the experiments include:
	\begin{itemize}
		\item We compare the classification accuracy and computation cost between different routing algorithms, based on the architecture described in~\cite{dynamicrouting}. 
		\item We also validate the effectiveness of our proposed architecture shown in Section \ref{sec:combining}, compared to the existing CapsNet architectures on several benchmark datasets.  %Based on our proposed architecture in Section \ref{sec:combining}, make a comparison with existing CapsNet architectures on several benchmark datasets.
		\item We investigate the ability of our proposed methods in maintaining the equivariance on image reconstruction tasks.
		\item Comparison of different methods on the performance of viewpoints invariance by experiments on smallNORB dataset.
	\end{itemize}
	\subsection{Comparison between Routing Algorithms}\label{sec:ex1}
	In this section, we compare our pairwise agreement algorithm with dynamic routing, based
	on the original CapsNet architecture~\cite{dynamicrouting} without the decoder. We use 3  routing iterations for dynamic routing, following~\cite{dynamicrouting},
	Experiments are conducted on 3 popular benchmark datasets: cropped-SVHN~\cite{SVHN}, Fashion-MNIST~\cite{Fashion-mnist},  CIFAR10~\cite{CIFAR10}. 
	%and run experiments on both dynamic routing (3 iterations) and our pairwise agreement algorithm to make a comparison. 
	We use Adam~\cite{adam} optimizer (beta1=0.9, beta2=0.999, epsilon=1e-7) with a learning rate of 0.001, batch size of 128. We do not use weight decay to simplify the discussion. 
	%Models are trained with a batch size of 128 and optimized by Adam~\cite{adam} with an initial learning rate of 0.001.  
	
	We train the models for 50 epochs without any data augmentation. Figure  \ref{fig:ori1} shows the training and test accuracy with respect to the training epochs. 
	We observe that our FM agreement has better optimization efficiency compared to the dynamic routing, over the three datasets. 
	Besides, our FM agreement obtains significantly better test accuracy than dynamic routing, which suggests its ability in improving the generalization of models.  
	% The result in Figure \ref{fig:ori1} shows that the accuracy on test sets is much better when using our FM agreement,  which demonstrates that our methods can effectively relieve from overfitting.
	
	We also investigate the experiment with data augmentation and run the experiments for 100 epochs. We use only random-crop on cropped-SVHN and Fashion-MNIST, while both random-crop and random-flip-left-right on CIFAR10. 
	The results are shown in Figure \ref{fig:ori2}. We also observe that our FM agreement obtains consistently better training performance than dynamic routing. 
	One interesting observation is that, compared to dynamic routing,  FM agreement has nearly the same test accuracy on cropped-SVHN,  slightly better on Fashion-MNIST, and much better on CIFAR10. 
	%From the results shown in Figure \ref{fig:ori2}, we can see that the test accuracy is almost the same on cropped-SVHN, and slightly better on Fashion-MNIST, but get much improvement on CIFAR10. 
	Based on this observation, we conjecture original dynamic routing is not robust on complex datasets with varied backgrounds, while our method is suitable at more complex datasets.
	
	\paragraph{Computational efficiency} We also compare the per batch inference time for models trained on CIFAR10 and Fashion-MNIST, which have different model sizes caused by the input sizes (32x32x3 and 28x28x1). 
	The time cost is computed on Nvidia GeForce GTX 1080ti GPU and Intel i7 CPU (3.5GHz).
	The results shown in Table \ref{tab:time compare} illustrate that the models using our FM agreement run faster than the dynamic routing both on GPU and CPU.
	\begin{table}
		\begin{center}
			\caption{The per-batch inference time. Dynamic routing (x) means using dynamic routing with x iterations.}
			\label{tab:time compare}
			\renewcommand\arraystretch{1.4}
			\begin{tabular}{c|cc|cc}
				\hline
				\multirow{2}{*}{Methods}&
				\multicolumn{2}{c|}{CIFAR10 (ms)}&
				\multicolumn{2}{c}{Fashion-MNIST (ms)}\\
				&GPU&CPU&GPU&CPU \\
				\hline
				Dynamic routing (1)   &5.79  &218.26  &4.43  &127.95 \\
				Dynamic routing (3)   &8.89  &274.58  &7.61  &157.69 \\
				FM agreement &\textbf{5.22} &\textbf{203.34} &\textbf{3.60} &\textbf{125.39} \\
				\hline
			\end{tabular}
		\end{center}
		% 		\vspace{-0.1in}
	\end{table}
	
	\begin{table}
		\begin{center}
			\caption{Performance comparison with existing CapsNet architectures on F-MNIST, CIFAR10 and SVHN. We report the classification accuracy with a form of ``mean$\pm$std'', computed on 3 random seeds.  }\label{tab:proposed capsnet}
			\renewcommand\arraystretch{1.2}
			\begin{tabular}{c|ccc}
				\hline
				\multirow{2}{*}{Model}&
				\multicolumn{3}{c}{Accuracy (\%)}\\
				&F-MNIST&CIFAR10&SVHN \\
				\hline
				Sabour \etal~\cite{dynamicrouting}   &93.60&89.40&95.70\\
				Hinton \etal~\cite{EMRouting}        &-&88.10&-\\
				Nair \etal~\cite{nair2018pushing}    &89.80&67.53&91.06\\
				HitNet~\cite{deliege2018hitnet}      &92.30&73.30&94.50\\
				DeepCaps~\cite{deepcaps}             &94.46&91.01&97.16\\
				Ours   &\textbf{94.70$\pm$0.17}&\textbf{93.20$\pm$0.24}&96.79$\pm$0.04\\
				\hline
			\end{tabular}
		\end{center}
		% 		\vspace{-0.1in}
	\end{table}
	
	\subsection{Performance on Proposed Architecture}\label{sec:ex2}
	In this experiment, we compare our proposed architecture described in Figure \ref{fig:complete arch} with the existing CapsNet architectures on   
	cropped-SVHN, Fashion-MNIST, CIFAR10. 
	We use 25 ResNet blocks and 3 capsule layers (with 32, 16, 10 capsules, respectively). 
	We follow the experimental setup described in the CIFAR10 experiment of ~\cite{resnet}. 
	We run the experiments with 3 random seeds and report the performance in Table~\ref{tab:proposed capsnet}. 
	
	From Table \ref{tab:proposed capsnet}, we can find that our model outperforms the other CapsNets on CIFAR10 with a significant margin, which further demonstrates the advantage of our model on complicated datasets. Our method also has slightly better performance on F-MNIST, while slightly worse performance on cropped-SVHN, compared to the state-of-the-art model DeepCaps~\cite{deepcaps}.
	Note that the parameters of our model are less than 1M, while DeepCaps~\cite{deepcaps} are 7.22M.
	We contribute  to that our proposed architecture adequately leverages the strength of ResNets~\cite{resnet} in accelerating training and improving the generalization.
	% 	reliable performance and easy to train.
	% 	Further, it also provides a feasible way to apply the capsule idea to many more tasks.

	\subsection{Reconstruction from the Pose Vector}\label{sec:ex3}
\begin{figure}[tb]
		\begin{center}
			\includegraphics[width=0.8\linewidth]{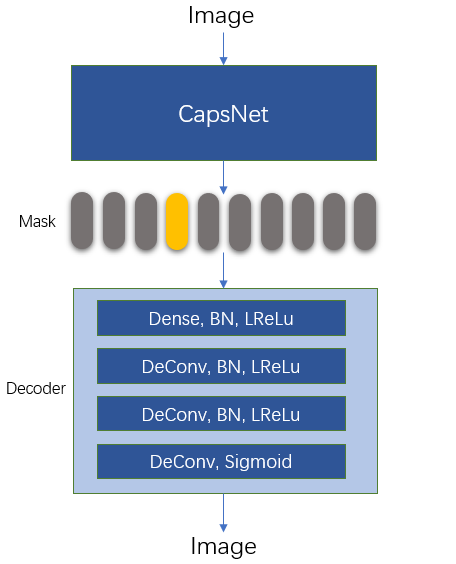}
		\end{center}
		% 		\vspace{-0.1in}
		\caption{The proposed model architecture which contains a decoder. The capsules from the CapsNet are masked out except for the ``correct'' one, then the decoder uses the ``correct'' capsule to reconstruct the input image.}
		\label{fig:decoder}
		\vspace{-0.1in}
	\end{figure}

	One of the key ideas in CapsNets is providing both equivariance and invariance properties.
	In this subsection, we investigate such properties by reconstructing the input image from the pose vector. 
	Figure~\ref{fig:decoder} describes the model architecture in this experiment, where we use the architecture in Figure~\ref{fig:complete arch} to generate the output capsules.
	We use the activations ($\hat{a}_j$ in Formula \ref{eqn:a_j}) of these capsules to compute the {\bf margin loss} (Formula~\ref{eqn:margin loss}). 
	The pose vectors $\hat{\mathbf{p}}_j$ (Algorithm \ref{alg:routing-by-fm}) of these capsules are all masked out except for the correct one (which is from the correct capsule indicated by the label during training, while from the capsule with the largest activation during the test). We then feed this vector into a decoder to reconstruct the input image. 
	Here, we build an efficient decoder with much fewer parameters by applying deconvolutional~\cite{deconvolutional} layers instead of fully-connected layers~\cite{dynamicrouting}, and only the correct $\hat{\mathbf{p}}_t \in \mathbb{R}^{1\times k}$ is fed into the decoder~\cite{deepcaps}. 
	With an additional reconstruction loss, the pose vector is encouraged to learn the properties such as rotation, translation, scale,~\etc  
	We use \emph{Adam}~\cite{adam} optimizer with the learning rate of 0.0001, and train the model with a batch size of 128.
	
	\subsubsection{Aff-NIST}\label{sec:ex3-1}
	We replicate the experiment in section 5.2 of the original paper~\cite{dynamicrouting} to evaluate the capability of our methods in terms of providing both invariance and equivariance. 
	We train our models on a padded and translated MNIST training set, in which each example is an MNIST digit placed randomly on a black background of $40\times 40$ pixels. 
	We then test it on the aff-NIST\footnote{\url{http://www.cs.toronto.edu/~tijmen/affNIST/}} dataset in which each example is an MNIST digit with a random small affine transformation, \emph{i.e.}, the models are never trained with affine transformations other than translation and any natural transformation seen in the standard MNIST. 
	Therefore, the results can indicate whether the models are robust on affine transformations. 
	
	We train models based on our architecture but using different routing algorithms (dynamic routing, EM routing, FM agreement) in the capsule layer and compare them on the test set of aff-NIST. 
	We run all the models 3 times and report their accuracy by ``mean$\pm$std'' in Table \ref{tab:aff result}.
	From the results, we can see that our method using the FM agreement achieves the best accuracy. 
	EM routing is vulnerable to numeric problems in our architecture according to our observation, so its performance is much worse than dynamic routing and FM agreement.
	Moreover, our model size is largely reduced, which contains about 199K parameters in the decoder and about 196K in the rest components. In contrast, the original CapsNet~\cite{dynamicrouting} contains about 2.2M parameters in the decoder and 11.2M in the rest components.
	\begin{table}
		\begin{center}
			\caption{Results on the test set of Aff-NIST. 
				% 			The top 2 rows are from the existing methods, and the bottom 3 rows are from our CapsNet architecture embedded with different routing algorithms. 
			}\label{tab:aff result}
			\renewcommand\arraystretch{1.4}
			\begin{tabular}{ccc}
				\hline
				Model&Params&Accuracy(\%) \\
				\hline
				Sabour \etal~\cite{dynamicrouting}  &13.40M&79.00 \\
				Hinton \etal~\cite{EMRouting}       &-&93.10 \\
				\hline
				Ours (Dynamic Routing)               &0.39K&92.06$\pm$0.71 \\
				Ours (EM Routing)                    &0.39K&78.29$\pm$2.04 \\
				Ours (FM agreement)                  &0.39K&\textbf{93.85$\pm$0.29} \\
				\hline
			\end{tabular}
		\end{center}
		% 	\vspace{-0.1in}
	\end{table}
	\begin{table}
		\begin{center}
			\caption{Sample reconstructions from the test set of Aff-NIST}\label{tab:aff recons}
			\renewcommand\arraystretch{1.2}
			\begin{tabular}
				{@{}p{8mm}@{\hskip 2.5mm}|@{\hskip 0.5mm}p{11mm}@{\hskip 0.5mm}p{11mm}@{\hskip 0.5mm}p{11mm}@{\hskip 0.25mm}|@{\hskip 0.25mm}p{11mm}@{\hskip 0.5mm}p{11mm}@{\hskip 0.5mm}p{11mm}@{\hskip 0.5mm}}
				&\multicolumn{3}{p{30mm}|@{\hskip 0.5mm}}{\centering expanded MNIST}&\multicolumn{3}{p{30mm}}{\centering aff-NIST}\\
				\hline
				\begin{minipage}[h]{9mm}
					Input \\ \\ Output
				\end{minipage}	
				&\begin{minipage}[h]{10mm}{\includegraphics[width=11mm, height=20mm]{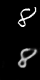}}\end{minipage}
				&\begin{minipage}[h]{10mm}{\includegraphics[width=11mm, height=20mm]{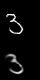}}\end{minipage}
				&\begin{minipage}[h]{10mm}{\includegraphics[width=11mm, height=20mm]{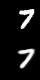}}\end{minipage}
				&\begin{minipage}[h]{10mm}{\includegraphics[width=11mm, height=20mm]{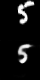}}\end{minipage}
				&\begin{minipage}[h]{10mm}{\includegraphics[width=11mm, height=20mm]{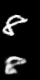}}\end{minipage}
				&\begin{minipage}[h]{10mm}{\includegraphics[width=11mm, height=20mm]{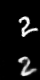}}\end{minipage}
			\end{tabular}
		\end{center}
	\end{table}
	\begin{table}
		\caption{ Sample reconstructions from the test set of Multi-MNIST. The top row shows the input images with overlapped digits, and the bottom row shows the reconstructed images with segmented digits generated by our CapsNet.}\label{tab:multi-mnist}
		\vspace{-0.1in}
		\begin{center}
			\renewcommand\arraystretch{1.2}
			\begin{tabular}{p{12.5mm}@{\hskip 0.5mm}|@{\hskip 0.5mm}p{12.5mm}@{\hskip 0.5mm}|@{\hskip 0.5mm}p{12.5mm}@{\hskip 0.5mm}|@{\hskip 0.5mm}p{12.5mm}@{\hskip 0.5mm}|@{\hskip 0.5mm}p{12.5mm}@{\hskip 0.5mm}|@{\hskip 0.5mm}p{12.5mm}}
				R:(5,6)&R:(2,1)&R:(3,9)&R:(9,0)&R:P:(5,2)&*R:(7,0)\\
				L:(5,6)&L:(2,1)&L:(3,9)&L:(9,0)&L:(5,9)&L:(7,8)\\
				\hline
				\begin{minipage}[h]{12.5mm}{\includegraphics[width=12.5mm, height=24mm]{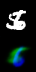}}\end{minipage}
				&\begin{minipage}[h]{12.5mm}{\includegraphics[width=12.5mm, height=24mm]{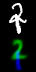}}\end{minipage}
				&\begin{minipage}[h]{12.5mm}{\includegraphics[width=12.5mm, height=24mm]{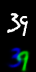}}\end{minipage}
				&\begin{minipage}[h]{12.5mm}{\includegraphics[width=12.5mm, height=24mm]{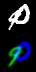}}\end{minipage}
				&\begin{minipage}[h]{12.5mm}{\includegraphics[width=12.5mm, height=24mm]{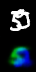}}\end{minipage}
				&\begin{minipage}[h]{12.5mm}{\includegraphics[width=12.5mm, height=24mm]{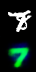}}\end{minipage}\\
			\end{tabular}
		\end{center}
		\vspace{0.1in}
	\end{table}
	
	As illustrated in Table \ref{tab:aff recons}, the reconstructions from pose vectors of expanded MNIST and aff-NIST successfully learned the crucial details of the input images such as thickness, translation, rotation, which demonstrates our methods can also provide equivariance for the input. 
	We believe it has the potential to build neural networks with better interpretability.
	
	\subsubsection{Multi-MNIST}\label{sec:ex3-2}
	We also replicate the experiment of multi-MNIST in~\cite{dynamicrouting}. The dataset in this experiment is created by merging each sample with a digit of other classes.
	All the samples are shifted up to four pixels randomly in each direction, resulting in a 36x36 image.
	For each digit image, we generate only 100 Multi-MNIST examples for the training and also 100 for testing, resulting in 6M images for the training set and 1M for the test set, which is a much smaller size compared to the original paper~\cite{dynamicrouting}. 
	
	We achieve a good result with an accuracy of 95.1\% on the training set and 94.5\% on the test set, which is similar to the result in paper~\cite{dynamicrouting}.
	However, the model is much smaller (457K parameters) compared to the original CapsNet (11.4M).
	
	We display some reconstructed images in Table \ref{tab:multi-mnist}, which shows the pose vectors generated by our model can segment the image into the 2 original digits just like the CapsNet in~\cite{dynamicrouting}, and we can see that the reconstructions still preserve the styles and positions from original inputs. 
	The top image shows the input, and the lower one shows the reconstructed digits overlayed in green and blue. 
	L:(l1; l2) represents the label for the two digits in the image, P:(p1; p2) represents the prediction for the two digits, and R:(r1; r2) represents the two digits used for reconstruction. 
	The penult column shows two examples with the wrong classification reconstructed from the prediction (P), which confuses digit `9' with digit `2'. 
	The last column with the (*) mark show the reconstruction from a digit that is neither the label nor prediction. 
	The other columns have correct classifications. 
	These results suggest that the model can find the best fit for all the input digits, including the ones that do not exist.
	
	\begin{table}
		\begin{center}
			\caption{Results on the test set of smallNORB.}\label{tab:small norb}
			\renewcommand\arraystretch{1.4}
			\begin{tabular}{ccc}
				\hline
				Model&Iteration&Accuracy(\%) \\
				\hline
				Dynamic Routing &1        &91.74$\pm0.86$ \\
				Dynamic Routing &2        &92.73$\pm0.62$ \\
				Dynamic Routing &3        &93.36$\pm0.69$ \\
				EM Routing      &1        &91.28$\pm1.02$ \\
				EM Routing      &2        &93.55$\pm0.75$ \\
				EM Routing      &3        &92.33$\pm0.64$ \\
				FM agreement    &-        &\textbf{93.60$\pm$0.53} \\
				\hline
			\end{tabular}
		\end{center}
		\vspace{0.2in}
	\end{table}
	
	\subsection{Viewpoints Invariance on SmallNORB}\label{sec:ex4}
	The smallNORB dataset~\cite{lecun2004learning} has gray-level stereo images of 5 classes of toys with every individual toy is pictured at 18 different azimuths (0-340), 9 elevations, and 6 lighting conditions.
	It is designed to be a pure shape recognition task without the context and color, so it becomes an important benchmark to evaluate the performance of viewpoints invariance.
	Hinton \etal~\cite{EMRouting} conducted experiments on smallNORB, and their proposed CapsNet with EM routing achieved impressive results.
	
	Based on our proposed architecture in Figure \ref{fig:complete arch}, we train models using different routing algorithms to make a fair comparison between our proposed FM agreement and other routing algorithms. 
	We train the model with SGD with a batch size of 64, a momentum of 0.9, and a weight decay of 0.0001. The training starts with a learning rate of 0.01 (which is divided by 2 for every 20 epochs) and is terminated at epoch 120.  
	We also run the experiments with 3 random seeds and show the results with a form of ``mean$\pm$std'' in Table \ref{tab:small norb}. 
	We again observe that our proposed FM agreement achieves the best accuracy, compared to the dynamic routing and EM routing.
	
	% 	While the convolutional capsule network in~\cite{EMRouting} using EM routing remains the state-of-the-art (98.20\%).
	% 	We conjecture that the complex implementation of EM routing makes it difficult to be adapted into different network architectures, and it still needs much more work to fine-tune the hyperparameters in EM routing to meet its best performance\footnote{Refer to the discussion of implementation for EM routing: https://openreview.net/forum?id=HJWLfGWRb }. 
	% 	Besides, we find that the model in~\cite{EMRouting} was trained on 8 sync GPUs, and for a day on smallNORB, 10 hours on MNIST and 2 days on CIFAR10 to achieve the results reported in ~\cite{EMRouting}, while our model only takes about 1 hour on a single GPU.
	\vspace{0.1in}
	\section{CONCLUSION}
	In this paper, we introduce an algorithm called FM agreement to improve the routing procedure in CapsNets.
	FM agreement is inspired by the FMs, which is widely used in the recommendation system. 
	We use FMs as a ``routing-by-agreement'' mechanism in the routing procedure, and conduct experiments to compare its performance with dynamic routing and EM routing. 
	The experiments illustrate that our method is of low complexity and outperforms the original dynamic routing and EM routing on several datasets. 
	We further propose a new CapsNet architecture combining both capsule layers and ResNet blocks, and experiments show it not only retains the strong discriminative performance of ResNets but also inherent the capability of CapsNets by providing both equivariance and invariance.
	
	Our future work includes investigating the performance of our CapsNet on more complicated datasets, and further exploring its equivariance and invariance quantitatively. 
	We believe this might be a potential direction for building deep neural networks with better robustness and interpretability.

	%\ack We would like to thank the referees for their comments, which
	%helped improve this paper considerably
	
	\bibliography{ecai,fm,face,capsnet,others}

\end{document}